\documentclass[%
 aip,
 amsmath,amssymb,
 reprint,%
]{revtex4-1}

\usepackage{graphicx}
\usepackage{dcolumn}
\usepackage{bm}

\usepackage[utf8]{inputenc}
\usepackage[T1]{fontenc}
\usepackage{mathptmx}
\usepackage{etoolbox}
\usepackage{nth}
\usepackage{todonotes}
\usepackage{verbatim}
\makeatletter
\def\@email#1#2{%
 \endgroup
 \patchcmd{\titleblock@produce}
  {\frontmatter@RRAPformat}
  {\frontmatter@RRAPformat{\produce@RRAP{*#1\href{mailto:#2}{#2}}}\frontmatter@RRAPformat}
  {}{}
}%
\makeatother
\begin{document}

\preprint{AIP/123-QED}

\title[High Dimensional Hybrid Reservoir Computing]{Predicting two-dimensional spatiotemporal chaotic patterns with optimized high-dimensional hybrid reservoir computing}
\author{T. Nakano}
 \email{tamon.nakano@dlr.de}
 \affiliation{Deutsches Zentrum für Luft- und Raumfahrt (DLR), Institut für KI Sicherheit, Rathausallee 12, 53757 Sankt Augustin, Germany} 
 
\author{S. Baur}%
 \affiliation{Deutsches Zentrum für Luft- und Raumfahrt (DLR), Institut für KI Sicherheit, Wilhelm-Runge-Str. 10, 89081 Ulm, Germany} 
 \affiliation{Department of Physics, Ludwig-Maximilians-Universität, Schellingstraße 4, 80799 Munich, Germany}

\author{C. Räth}
 \affiliation{Deutsches Zentrum für Luft- und Raumfahrt (DLR), Institut für KI Sicherheit, Wilhelm-Runge-Str. 10, 89081 Ulm, Germany} 
 \affiliation{Department of Physics, Ludwig-Maximilians-Universität, Schellingstraße 4, 80799 Munich, Germany}

\date{\today}

\begin{abstract}
As an alternative approach for predicting complex dynamical systems where physics-based models are no longer reliable, reservoir computing (RC) has gained popularity. The hybrid approach is considered an interesting option for improving the prediction performance of RC. The idea is to combine a knowledge-based model (KBM) to support the fully data-driven RC prediction. 
There are three types of hybridization for RC, namely full hybrid (FH), input hybrid (IH) and output hybrid (OH), where it was shown that the latter one is superior in terms of the accuracy and the robustness for the prediction of low-dimensional chaotic systems.
Here, we extend the formalism to the prediction of spatiotemporal patterns in two dimensions. To overcome the curse of dimensionality for this very high-dimensional case we employ the local states ansatz, where only a few locally adjacent time series are utilized for the RC-based prediction.        
Using simulation data from the Barkley
model describing chaotic electrical wave propagation
in cardiac tissue, we outline the formalism of high-dimensional hybrid RC and assess the performance of the different hybridization schemes.
We find that all three methods (FH, IH and OH) perform better than reservoir only, where improvements are small when the model is very inaccurate. For small model errors and small reservoirs FH and OH perform nearly equally well and better than IH. Given the smaller CPU needs for OH and especially the better interpretability of it, OH is to be favored.
For large reservoirs the performance of OH drops below that of FH and IH.  
Generally, it may be advisable to test the three setups for a given application and select the best suited one that optimizes between the counteracting factors of prediction performance and CPU needs. 

\end{abstract}

\maketitle

\begin{quotation}
The prediction of high-dimensional spatiotemporal patterns emerging from nonlinear complex systems like turbulent flows, excitable media and earth systems, is an essential task in vatious fields of science. Especially in physics and engineering approximate or reduced order models of the underlying dynamical phenomenon are available in many cases. While these models are still too inaccurate to be used for precise predictions, they may be combined with fully data-driven, AI-based methods to allow for precise predictions. In this paper we extend these hybridization techniques to high-dimensional, spatially extended dynamical systems using reservoir computing (RC) as AI-method. We thoroughly test different setups for hybrid RC and discuss the respective performance gain. Our results suggest that the prediction of spatiotemporal patterns is significantly improved by hybrid RC allowing for a whole new range of conceivable applications.\\

\end{quotation}

\section{\label{sec:introduction}Introduction}

Reservoir computing (RC)~\cite{jaeger2001echo, maass2002real, jaeger2004harnessing} has gained popularity as an alternative solution to predict complex dynamical systems where physics-based models are no more reliable, because it combines superior forecasting results with little computational needs~\cite{chattopadhyay20,bompas20,shahi22}. RC is by default a fully data-driven approach and is expected to learn the dynamics of the underlying system from the data. 
The hybrid approach~\cite{pathak2018hybrid} is considered as an interesting option to improve the prediction. The idea is to combine a knowledge-based model (KBM, e.g. an imperfect governing equation) as a support to the fully data-driven prediction by the reservoir. 
The hybrid approach can even be pursued if no governing equations are available. In this case one can derive proxies of the governing equations from the data with e.g. SINDy~\cite{brunton16} or a causality analysis~\cite{ma22} and continue with hybrid RC. Also this approach leads to improvements in the prediction performance especially when the reservoir parameters are not optimized~\cite{koester23}.\\   
The combination of data-driven and model-based elements in
hybrid reservoir computing can be done at the input or output layer of the RC or both of them. The three setups are respectively called, input hybrid (IH), output hybrid (OH) and full-hybrid (FH). 
Some studies have already been performed, for example, on the input-hybrid by Shahi et al.~\cite{shahi2021long}, full-hybrid by Pathak et al.~\cite{pathak2018hybrid}  
and on the output-hybrid by Doan et al.~\cite{doan2019physics}$^,$~\cite{doan2021short}. 
Duncan et al.~\cite{duncan2023optimizing} analyzed and compared the different setups in a systematic manner and showed more recently the superiority of the output-hybrid setup in terms of in terms of accuracy, robustness and interpretabiliy of the results for a set of three-dimensional chaotic model systems.\\

The capability of RC to predict extended, high-dimensional spatiotemporally chaotic systems is also of great interest in complex systems research. 
The predicted systems in the above mentioned studies of hybrid RC have low dimensionalities ranging form $3$ to $6$. 
To treat high-dimensional systems one possible solution is to use a large reservoir (i.e. a large number of reservoir nodes), but this could make the training unfeasible. This is called “curse of dimensionality”, as one can also observe in many other machine learning methods. 
Parlitz et al. ~\cite{parlitz2000prediction} suggested a parallel prediction approach based on local states (LS), where only a few locally adjacent time series are utilized for the prediction. This procedure is repeated for all input time series to be predicted. 
Pathak et al. applied this LS-approach to RC~\cite{ pathak2018model} and showed its efficiency for a one-dimensional system (Kuramoto-Sivashinsky) consisting of $512$ time series. 
Zimmermann et al. ~\cite{zimmermann2018observing} applied RC with LS to (two-dimensional) spatially extended systems, namely models describing excitable media
(Barkley and the Bueno-Orovio-Cherry-Fenton model) and 
performed a prediction between the variables of the system.
The two variables $U(t)$ and $V(t)$ of the system were cross-predicted, i.e. predicted from each other. Their prediction can thus rather be considered as a mapping or a reconstruction at each time step than a true temporal prediction in the usual sense.\\
Wikner et al. ~\cite{wikner2020combining} proposed an approach that combines hybrid RC with local states and called it Combined Hybrid-Parallel Prediction (CHyPP). In the study, CHyPP was compared to the non-hybrid RC and the non-parallel hybrid RC. CHyPP showed a better prediction quality and required smaller dataset for the training compared to the other two approaches. 
However, the studied system was one-dimensional (again Kuramoto-Sivashinsky) with $1024$ dimensions at highest and only output hybrid was considered. 

There is a high expectation for RC and hybrid RC to be capable of predicting spatially extended, much higher dimensional systems for which only proxies of a reliable physics-based model are available. 
Atmospheric models and many phenomena in fluid dynamics are examples. 


In this paper we extend the formalism of hybrid Reservoir Computing to the prediction of high-dimensional, spatially extended two-dimensional data sets by using Local States.
We calculate and validate the predictions with IH, OH and FH and systematically study the results for different sizes of the reservoir and varying model mismatches and compare them with the reservoir only results.

\section{Barkley Model}
\label{sec:barkley_model}
As outlined in Zimmermann et al. ~\cite{zimmermann2018observing} we use the (cubic) Barkley model describing chaotic electrical wave propagation in cardiac tissue as two-dimensional model system.
It is a non-linear chaotic system which is described by partial differential equations with a \nth{2} order diffusion term,

\begin{equation}
\begin{aligned}
& \frac{\partial U}{\partial t}=D \cdot \nabla^2 U+\frac{1}{\epsilon}U(1-U)\left(U-\frac{V+b}{a}\right) \\
& \frac{\partial V}{\partial t}=U^3-V.
\label{eq:barkley}
\end{aligned}
\end{equation}

It describes the dynamics of two coupled variables $U(t)$ and $V(t)$ that depend on the four coefficients $\epsilon$, $a$, $b$, $D$. 
The numerical solution of the two partial differential equations has been performed with an \nth{1} order Euler method for the time integration of $\Delta t=0.01$. A grid of $80\times80$ points with $\Delta x=0.1$ constantly all over the grid was used. Thus the two-dimensional data sets consists of $6400$ simultaneously evolving time series. 
A second order spatial discretization was used for the approximation of the Laplace operator. The boundary condition was set to the no-flux condition. The coefficients were set to be $D=0.02$, $a=0.75$, $b=0.06$ and $\epsilon=0.08$. The source code for the simulation is available\cite{zimgit} by courtesy of Zimmermann and Parlitz. 
Fig.\ref{barkley_model} shows the evolved $U(t)$ at a randomly captured moment in the simulation. 

It is known from previous studies~\cite{baur21} that adding noise to input time series data improves the 
long-term prediction at the cost of the short-term 
prediction quality. Thus a normally distributed noise with standard deviation of $\sigma_\text{SD,noise}$ was added for the training and the synchronization of the training. Given that $\sigma_\text{SD,input}$ is the standard deviation of the input time series, we set to $\sigma_\text{SD,noise}=\alpha\sigma_\text{SD,input}$, when the noise-ratio $\alpha$ is a hyper-parameter.

In the hybrid approach, the $\epsilon$-model was employed. This was originally suggested by Pathak et al.~\cite{pathak2018hybrid}. The idea is to vary a parameter of the governing equations of the system and then use it as the KBM. Here the same numerical integration method as for the ground truth (first order Euler method in our case) was employed. By doing so, an imperfect predictor is artificially created. In our case, we multiplied the coefficient $\epsilon$ in the Barkely model by the factor of $(1+e)$ which gives $\epsilon_{e}=\epsilon(1+e)$. $e$ is called a model error. $\epsilon$ in the equation $\label{eq:barkley}$ will be replaced by $\epsilon_{e}$ for the KBM. Although the model error is often denoted by $\epsilon$ as the this approach named "$\epsilon$-model", we denote here the model error $e$ to avoid the confusion with $\epsilon$ in the equations of the Barkley model.

\section{Reservoir Computing}
\label{sec:reservoir}
Let $\bm{u}$ denote the $u$-dimensional time series used for the training and the synchronization. During the prediction, the previous prediction $\bm{y}_\text{r}(t-\Delta t)$ is used as the input for the next prediction. Let $\tilde{\bm{u}}(t)$ denote all these inputs time series. In the case of a classical, non-hybrid RC, a $x_\text{dim}$-dimensional input vector $\bm{x}(t)$ for the reservoir is given as $\bm{x}(t)=\tilde{\bm{u}}(t)$.
In the case of hybrid RC the input vector becomes a function of the input time series, $\bm{x}(t) = f_\text{inp}(\tilde{\bm{u}}(t))$. $f_\text{inp}$ depends on the type of hybrid RC and is further specified below.

RC is a type of recurrent neural networks (RNNs). The uniqueness of the RC is that the weights in the reservoir and in the input layer are fixed during the training. The reservoir is a sparse random network consisting of $r_\text{dim}$ nodes with the average node degree $\kappa$. The random connections between the nodes and their strengths (weights) are described by the $r_\text{dim} \times r_\text{dim}$-dimensional adjacency matrix $\mathbf{A}$.  The weights are scaled to have a spectral radius $\rho$. The input time series $\bm{x}(t)$ are connected to the nodes of the network by the input matrix $\mathbf{{W}}_\text{in}$ which is a sparse $r_\text{dim}\times x_\text{dim}$ dimensional matrix.  Concretely, the input data $\bm{x}(t)$ are used to define the reservoir state $\bm{r}(t+1)$ with the one in the previous time step $\bm{r}(t)$,
$$\bm{r}(t+\Delta t)=\tanh \left[\mathbf{A} \bm{r}(t)+\mathbf{W}_{\text{in}} \bm{x}(t)\right] .$$


$\mathbf{A}$ and $\mathbf{{W}}_\text{in}$ are initialized once before the training and then kept fixed. In order to break detrimental symmetries in the reservoir equations~\cite{herteux20} we introduce a non-linear transformation of the reservoir state $\tilde{\bm{r}}(t)$, $$\tilde{\bm{r}}=\left[ \bm{r}, \bm{r}^{2} \right]^{T}=\left[ r_1, r_2, ..., r_{\text{dim}}, r^{2}_{1}, r^{2}_{2}, ..., r^{2}_{\text{dim}} \right]^{T}.$$

Then the reservoir state $\tilde{\bm{r}}(t)$ is again transformed to $h_{dim}$-dimensional vector $\bm{h}(t)$ that can depend on the transformed reservoir response 
$\tilde{\bm{r}}(t)$ and the input data $ \tilde{\bm{u}}(t)$,
$\bm{h}(t) = f_\text{out}(\tilde{\bm{r}}(t), \tilde{\bm{u}}(t))$. Again in the classical RC $\bm{h}(t)$ is simply given by $\bm{h}(t)=\tilde{\bm{r}}(t)$, whereas in the hybrid case
the contributions of the KBM are added via a proper definition of $f_\text{out}$. 

The output matrix $\mathbf{W}_\text{out}$ connects  $f_\text{out}$ to the target output $\bm{y}_\text{t}(t)$. $\mathbf{W}_\text{out}$ is determined in the training by a ridge regression,

$$\mathbf{W}_{\text{out}}=\min _{\mathbf{W}_{\text{out}}}\left[\left\|\mathbf{W}_{\text{out}} {\bm{h}}(t)-\bm{y}_\text{t}(t)\right\|+\beta\left\|\mathbf{W}_{\text{out}}\right\|\right] \text {. }$$
$\beta$ is the regularization term to avoid an over-fitting. Once trained, the output $\bm{y}_\text{r}$ can be given by $\bm{y}_\text{r}=\textbf{W}_\text{out}{\bm{h}}(t)$. The prediction is done progressively using the prediction of the previous time step.

$$\bm{r}(t+\Delta t)=\tanh \left[\mathbf{A} \bm{r}(t)+\mathbf{W}_{\text{in}} \bm{x}_{\text{pred}}(t)\right].$$

\section{Hybrid approach}
\label{sec:hybrid_approach}
We employ three different hybrid approaches as outlined in the study by Duncan et al.~\cite{duncan2023optimizing}. The KBM can be represented as a function being applied on an input time series $\bm{u}$ and producing an imperfect next-step prediction,

$$
K(\bm{u}) \approx \bm{u}(t+\Delta t).
$$

In the input hybrid (IH) method, the reservoir input $\bm{x}(t)$ is given as the concatenation of the input $\tilde{\bm{u}}(t)$ and the KBM output $K(\tilde{\bm{u}}(t))$, $$ \bm{x}(t)=f_{\text{inp}}(\tilde{\bm{u}}(t))=\left[\begin{array}{c}
\tilde{\bm{u}}(t) \\
K(\tilde{\bm{u}}(t))
\end{array}\right] .
$$
This extends the input dimension of a classical (non-hybrid) RC, $x_{\text{dim}}=u_{\text{dim}}$, to $x_{\text{dim}}=u_{\text{dim}}+K_{\text{dim}}$. Therefore the input matrix $\mathbf{W}_{\text{in}}$ is now a $r_{\text{dim}} \times\left(u_{\text{dim}}+K_{\text{dim}}\right)$ matrix. At the output layer only the reservoir response and its square is used, i.e. $\bm{h}(t)=\tilde{\bm{r}}(t)$\\ 

In the output hybrid (OH) method, the KBM prediction $K(\tilde{\bm{u}}(t))$ is only
fed into the output layer, i.e.
$\bm{h}(t)=\bm{r}(t)$ in the classical (non-hybrid) RC is transformed to 
$$ \bm{h}(t)=f_{\text{out}}(\bm{r}(t), \tilde{\bm{u}}(t))=\left[\begin{array}{c}
\bm{r}(t) \\
K(\tilde{\bm{u}}(t))
\end{array}\right] .
$$
The output matrix $\mathbf{W}_{\text{out}}$ now has the shape of $u_{\text{dim}}\times\left(r_{\text{dim}}+K_{\text{dim}}\right)$.
Conversely, the input layer is left unchanged in the OH method, i.e. $\bm{x}(t)=\tilde{\bm{u}}(t)$.\\  

\begin{figure}
\begin{minipage}[c]{0.49\linewidth}
\includegraphics[width=\linewidth]{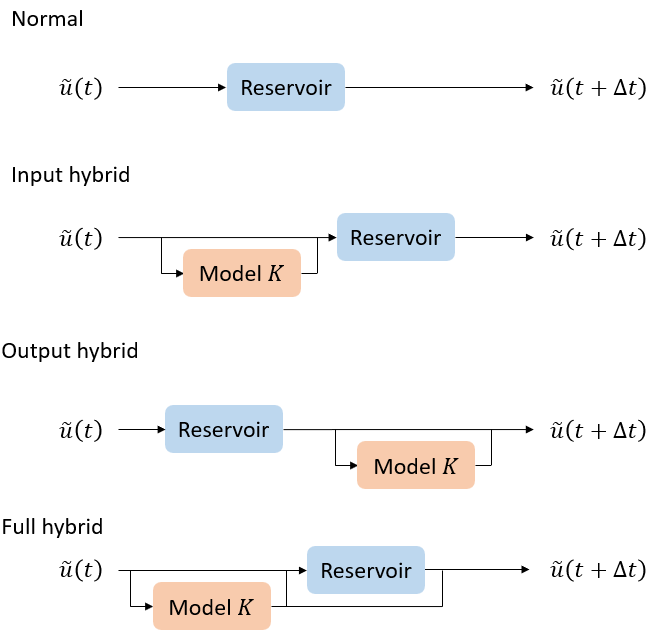}
\caption{Schematic illustration of the different hybrid methods considered in this study}
\label{hybrid}
\end{minipage}
\hfill
\begin{minipage}[c]{0.49\linewidth}
\includegraphics[width=\linewidth]{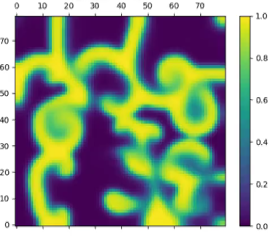}
\caption{$U(t)$ given by the simulation with the Euler method at a randomly chosen instant}
\label{barkley_model}
\end{minipage}%
\end{figure}

The full hybrid (FH) method is simply a method where both the input- and output-hybrid are simultaneously employed. 
Fig.\ref{hybrid} schematically illustrates the three hybrid RC methods.

In our example the input time series have the dimension of $\tilde{\bm{u}}(t)=80\times80\times2$ per time step, which is the multiplication of the number of the grid points and the number of the variables. $K_{\text{dim}}$ has also $80\times80\times2$ dimensions.

\section{Local States}
\label{sec:local_states}
Performing the training and the prediction on a $80\times80\times2$ data-set at each time step can be computationally expensive and one can get trapped in the “curse of dimensionality”, as previously mentioned. In a preliminary study, we attempted to perform the training and prediction with one single RC which covers the whole $80\times80\times2$ domain. This approach requires a significantly long dataset for the training. The required computational memory is also huge. With limited computational resource and limited data length, we got very poor prediction performance. 
These results further suggest the use of Local States (LS) for the prediction of high-dimensional extended systems.
The single RC approach won't be further considered in this paper. 

\begin{figure}
\centering
\includegraphics[width=0.25\textwidth]{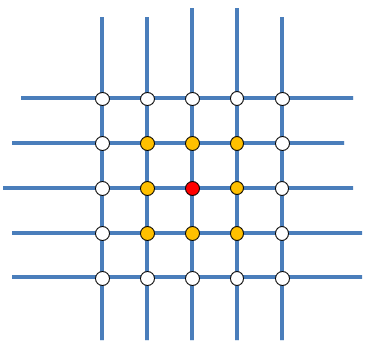}
\caption{Local States: the training and the prediction are done for the red point taking only the information of the red point itself and the orange points.} \label{local_states}
\end{figure}

LS is a parallel prediction approach in which only the local behavior of the system is considered by taking into account only neighboring grid points. To predict the variables $U(t)$ and $V(t)$ at the grid point of $(i, j)$, we use the information of the neighbouring points in a $\sigma \times \sigma$ sized square, including $(i, j)$ itself. Fig.\ref{local_states} shows the concept.
Physically this approach is justified, when the interaction lengths in the dynamical system under study is limited and covered by the local neighborhood.
$\sigma=3$ has been chosen from the results of a hyper-parameter study. This gives a $\sigma^{2}\times2$ dimensional vector as input, taking both variables $U(t)$ and $V(t)$. In our study, a reservoir has been initialized, trained and performed the prediction at each point in the domain $(i, j)$. All these procedures have been done separately for each point on the grid. This means that $80\times80$ independent reservoirs have been used. 

\section{Valid Time}
\label{sec:valid_time}
We evaluated the quality of the prediction by the valid time. The valid time is the time during which one can have a prediction with an error smaller than a certain acceptable level. The valid time $t_\text{v}$ is given as the duration before the normalized time-dependent error $e(t)$,
$$
e(t) = \frac{\| y(t) - y_\text{r}(t) \|}{\langle \| y(t)\|^2 \rangle^{1/2}}
$$
exceeds a threshold value $e_\text{max}$. In this study
$e_\text{max}$ was set to 0.2.

\section{Ensemble experiment method}
\label{sec:ensemble}
The quality of the prediction by a RC is, as many other machine learning methods, dependent on the specific time series sections used for the training and the prediction. Therefore it is important to asses the performance of the prediction in a statistically meaningful way, such as an ensemble experiment. This method was employed in Figs.\ref{reservoir_rdim}, \ref{model_error_100} and \ref{model_error_500} in Sec. \ref{sec:prediction}.

\begin{figure}
\centering
\includegraphics[width=0.25\textwidth]{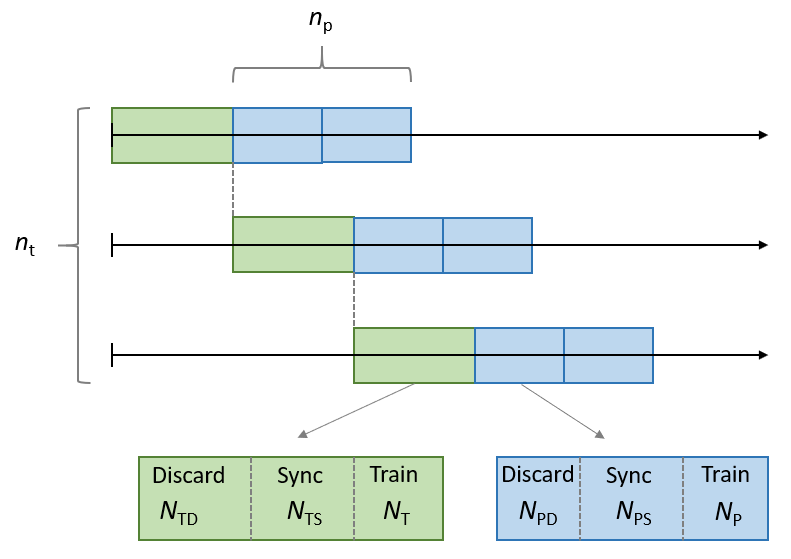}
\caption{Training and prediction sections in the ensemble experiment. $n_\text{T}=2, n_\text{P}=3$ in this study.} \label{ensemble_experiment}
\end{figure}

In total $n_\text{T}$ trainings have been performed. Each training-section $i_\text{T}$ is composed of $N_\text{TD} + N_\text{TS} + N_\text{T}$ time steps; $N_\text{TS}$ steps of train-synchronization, $N_\text{T}$ of training. The first $N_\text{TD}$ steps of each section are discarded to avoid the influence from the last section. For a each training $i_\text{T}$, $n_\text{P}$ times prediction has been done. Each prediction section consists of $N_\text{PD} + N_\text{PS} + N_\text{P}$ time steps, where the first $N_\text{PD}$ time steps are again discarded. The example in Fig.\ref{ensemble_experiment} shows the case of $n_\text{T}=2, n_\text{P}=3$.

\section{Hyper-parameter Study}
\label{sec:hyperparameter}
A hyper-parameter study has been conducted to improve the prediction quality for the Barkley model introduced in Sec.\ref{sec:barkley_model}. The evaluation was done by a single training and a single prediction. The training has been done with 10.000 steps. The prediction quality was measured by the valid time. Table.\ref{tab:hyper_parameter} summarizes the examined values and the obtained optimal values. When a parameter was examined (varied), the other parameters were fixed to the initial value. In the following section, the optimal values were used unless otherwise described. The adjacency matrix $A$ and the input matrix $W_\text{in}$ were randomly initialized only once at the beginning and fixed during the entire study.

\begin{table}[h!]
   \caption{Hyper-parameter study} 
   \label{tab:hyper_parameter}
   \small
   \centering
   \resizebox{0.49\textwidth}{!}{
   \begin{tabular}{l|l|l|l}
   \text{Parameters} & \text{Examined values} & \text{Initial value} & \text{Optimal value} \\
   \toprule
   $r_\text{dim}$ & $\in$ $\{200, 400, 500, 600\}$ & $500$ & $400$ \\
   $\rho$ & $\in$ $\{0.1, 0.3, 0.5, 0.6, 0.7, 0.8, 1.0, 1.2, 1.5\}$ & $1.0$ & $0.5$ \\   
   $\sigma$ & $\in$ $\{3, 5, 7\}$ & $5$ & $3$ \\
   $\alpha$ & $\in$ $10^{-x}, x=\{4, 5, 6\}$ & $10^{-4}$ & $10^{-6}$ \\ 
   $\beta$ & $\in$ $10^{-x}, x=\{2, 3, 4, 5, 6, 7, 8\}$ & $10^{-6}$ & $10^{-6}$ \\
   \end{tabular}}
\end{table}

\section{Prediction Results}
\label{sec:prediction}
The Barkley model has been simulated for 40,400 steps. The first $T_\text{transi}=2,000$ steps were discarded to avoid the transition regime in the simulation. The next $T_\text{train,sync}=200$ steps were used for the synchronization for the training, and the next $T_\text{train}=30,000$ steps were used for the training. Again $T_\text{pred,sync}=200$ steps were used for the synchronization for the prediction, and finally $T_\text{pred}=8,000$ steps are employed for the prediction. The computation time for the training plus the prediction is typically around 24 hours with a CPU of the 12th Gen Intel(R) Core(TM) i9-12900K. The reservoir dimension $r_\text{dim}$ and the training $T_\text{training}$ and prediction steps $T_\text{pred}$ are the major factors for the computation time.

\begin{figure}[h!]
  \begin{minipage}[b]{\linewidth}
  \includegraphics[width=\linewidth]{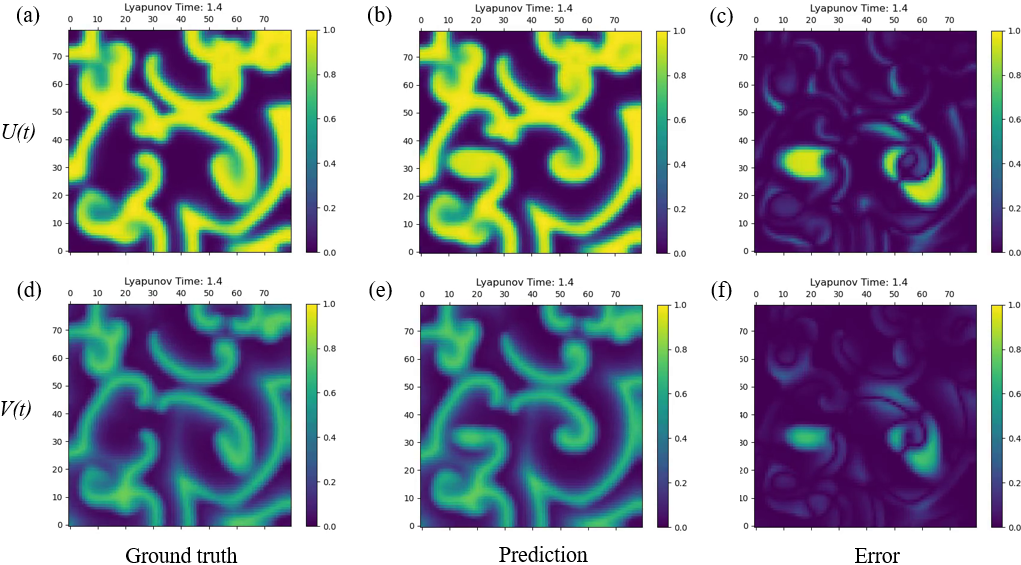}
  \medskip\footnotesize\centering instant A, Lyapunov Time $t\lambda_\text{max}=1.4$ 
  \end{minipage}
  \vfill
  \begin{minipage}[b]{\linewidth}
  \includegraphics[width=\linewidth]{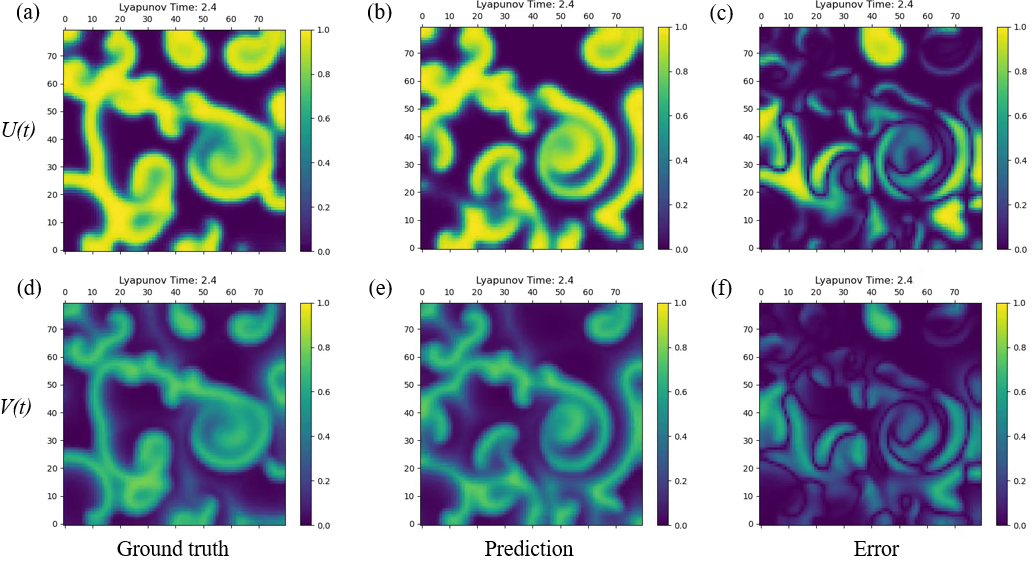}
  \medskip\footnotesize\centering instant B, Lyapunov Time $t\lambda_\text{max}=2.4$ 
  \end{minipage}
  \caption{Prediction by the reservoir-only, $U(t)$: (a)-(c), $V(t)$: (d)-(f), ground truth, prediction and error $e(t)=|\bm{y}_\text{t}(t) - \bm{y}_\text{r}(t)|$, from left to right, where $\bm{y}_\text{t}(t)$ is the ground truth, $\bm{y}_\text{r}(t)$ is the prediction.}
  \label{reservoir_only}
\end{figure}

Fig.\ref{reservoir_only} shows the reservoir-only prediction (non-hybrid approach), and the comparison to the ground truth at a randomly chosen moment, the instant A (Lyapunov Time $t\lambda_\text{max}=1.4$) and the instant B ($t\lambda_\text{max}=2.4$). At the instance A, at least qualitatively the prediction captures well the dynamics of the system. The prediction error is obviously larger at the instant B than A.

\begin{figure}[h!]
  \begin{minipage}[b]{\linewidth}
  \includegraphics[width=\linewidth]{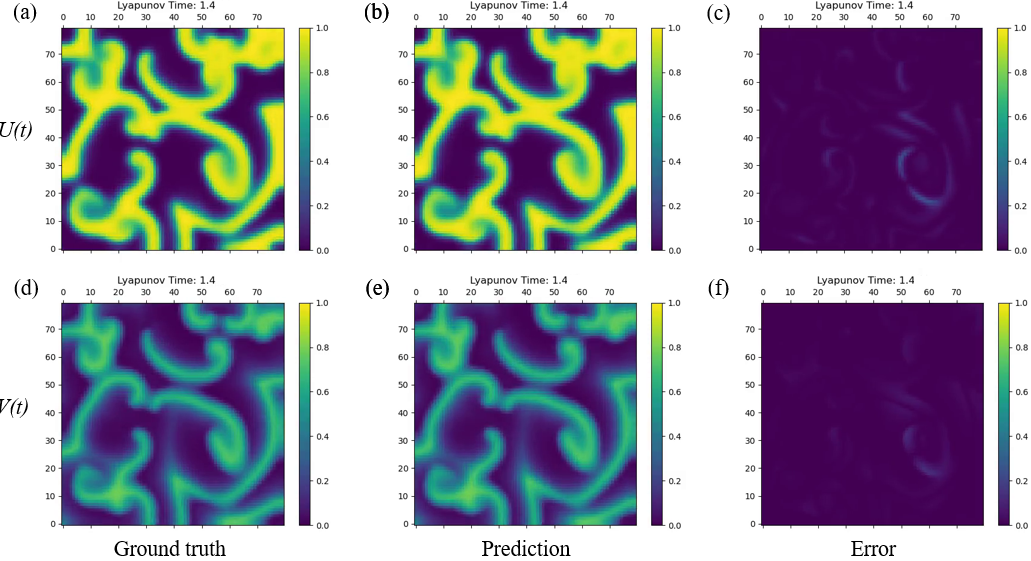}
  \medskip\footnotesize\centering instant A, Lyapunov Time $t\lambda_\text{max}=1.4$ 
  \end{minipage}
  \vfill
  \begin{minipage}[b]{\linewidth}
  \includegraphics[width=\linewidth]{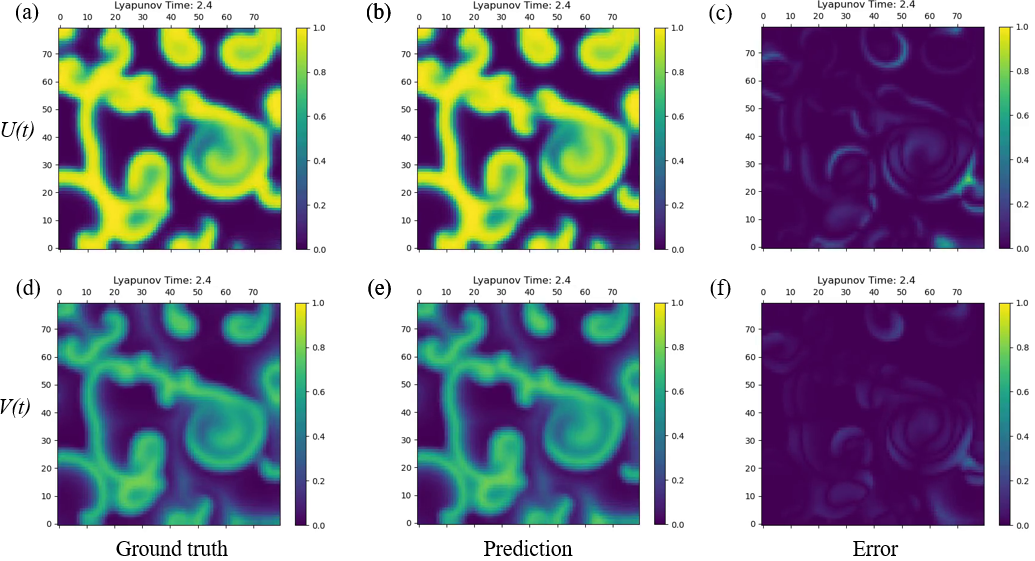}
  \medskip\footnotesize\centering instant B, Lyapunov Time $t\lambda_\text{max}=2.4$ 
  \end{minipage}
  \caption{Same as Fig. \ref{reservoir_only} but prediction by the OH.}  
  \label{output}
\end{figure}

Fig.\ref{output} shows the prediction by the OH and the comparison to the ground truth at the same randomly chosen moment as before, i.e. at instant A with $t\lambda_\text{max}=1.4$ and at instant B with $t\lambda_\text{max}=2.4$. The model error was set to $e=0.1$ and $r_\text{dim}=400$. In both instants, at least qualitatively the prediction captures again well the dynamics of the system. As expected the prediction error is larger at the instant B. At both instances, the prediction error is alleviated compared to the reservoir-only case in Fig.\ref{reservoir_only}, which implies that the prediction was significantly improved with the help of the OH.

\begin{figure}
\includegraphics[width=\linewidth]{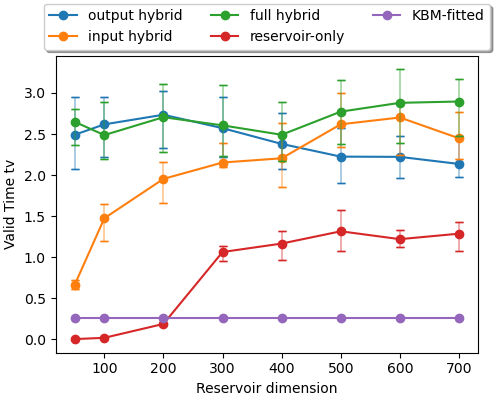}
\caption{Reservoir dimension $r_{dim}$ vs valid time with $e=0.1$. The valid time is estimated by calculating the median of all the $n_\text{T}\times n_\text{P}$ predictions and displayed with the corresponding standard deviations (lower/higher quartile) across each dimension. The KBM-fitted prediction is also displayed for the comparison.} 
\label{reservoir_rdim}
\end{figure}

\begin{figure}
\includegraphics[width=\linewidth]{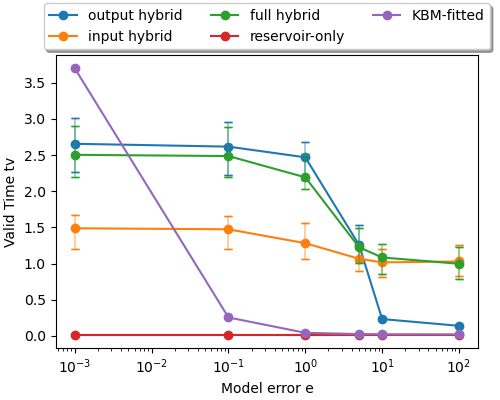}
\caption{Model error $e$ vs valid time for the reservoir dimension $r_{dim}=100$. The valid time is estimated by calculating the median of all the $n_\text{T}\times n_\text{P}$ predictions and displayed with the corresponding standard deviations (lower/higher quartile) across each dimension. The KBM-fitted prediction is also displayed for the comparison.}
\label{model_error_100}
\end{figure}

\begin{figure}
\includegraphics[width=\linewidth]{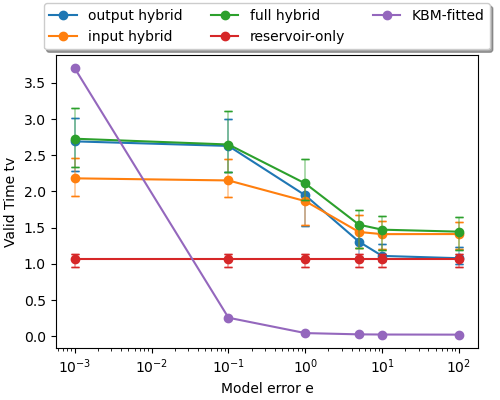}
\caption{Same as Fig. \ref{model_error_100} but for the reservoir dimension $r_{dim}=300$.}
\label{model_error_300}
\end{figure}

\begin{figure}
\includegraphics[width=\linewidth]{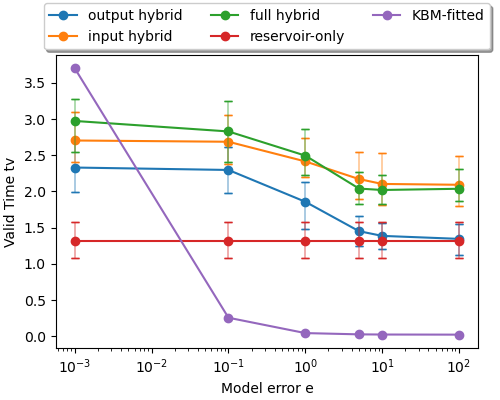}
\caption{Same as Fig. \ref{model_error_100} but for the reservoir dimension $r_{dim}=500$.}
\label{model_error_500}
\end{figure}

In Fig.\ref{reservoir_rdim}-\ref{model_error_500}, the prediction performance in terms of the model error $e$ and the reservoir dimension $r_\text{dim}$ was investigated. The ensemble experiment has been done with $n_\text{T}=3, n_\text{P}=6$, 18 experiments in total (see Sec.\ref{sec:ensemble} for the definition). As in Fig.\ref{reservoir_only} and Fig.\ref{output}, the data lengths were set to be $T_\text{transi}=2,000$, $T_\text{train,sync}=200$, $T_\text{train}=30,000$, $T_\text{pred,sync}=200$, $T_\text{pred}=8,000$. The adjacency matrix $A$ and the input matrix $W_\text{in}$ has been initialized only once at the first training-section $i_\text{T}=1$ and have been kept for all the rest of the sections.

Fig.\ref{reservoir_rdim} shows the valid time $t_\text{v}$ of the different hybrid methods versus the reservoir dimension $r_\text{dim}$. The reservoir-only case actually performs worse than the KBM-fitted prediction at lower dimensions but as the dimension increases its performance is improved. The IH improves the valid time around 1 compared to the reservoir-only case at the tested range of the dimension $r_\text{dim}$. The OH and the FH perform already well in the case of smaller reservoirs. The prediction performances decreases for OH with increasing the dimension $r_\text{dim}$ of the reservoir. As explained in Sec.\ref{sec:hybrid_approach}, the output matrix $\mathbf{W}_{\text{out}}$ has $u_{\text{dim}}\times\left(r_{\text{dim}}+K_{\text{dim}}\right)$ dimensions in total, $u_{\text{dim}}\times r_{\text{dim}}$ for the reservoir part and $u_{\text{dim}}\times K_{\text{dim}}$ for the KBM part. This indicates that the increase of the reservoir dimension $r_\text{dim}$ makes the contribution of the KBM smaller in the OH. This is why the increase of the reservoir dimension $r_\text{dim}$ in the OH actually lowers the prediction performance. In case of the IH, the input matrix $\mathbf{W}_{\text{in}}$ has $r_{\text{dim}} \times\left(u_{\text{dim}}+K_{\text{dim}}\right)=r_{\text{dim}}\times36$ dimensions in total which can be split up inot $r_{\text{dim}} \times u_{\text{dim}}=r_{\text{dim}}\times18$ for the reservoir part $r_{\text{dim}} and \times K_{\text{dim}}=r_{\text{dim}}\times18$ for the KBM part. This means that the increase of the reservoir dimension $r_\text{dim}$ does not make the contribution of the KBM smaller. This is why the IH shows rather better performance with a larger dimensions. The FH seems to take the advantages of both the IH and the OH, where it is noteworthy that neither at low dimensions nor at high dimensions of the reservoir FH strongly exceeds the prediction performance of OH or IH respectively.  

In Fig.\ref{model_error_100}-\ref{model_error_500}, the valid time of the different hybrid methods as a function of the model error $e$ is compared, with the reservoir dimension $r_\text{dim}=100, 300$ and $500$ respectively. All hybrid methods outperform the reservoir-only at smaller model errors $e$. 
This is especially true for the case of a small reservoir $r_\text{dim}=100$,  where reservoir-only completely fails to make accurate predictions.
As expected, the hybrid methods get more inaccurate as the model error $e$ increases. For the OH, when the model error $e$ is large, the correction by the reservoir part at the output matrix $\textbf{W}_\text{out}$ is not enough and it simply tries to get rid of the contribution from the KBM. No contribution from the KBM means actually nothing else than a non-hybrid reservoir. This is why at large errors such as $e=10$ or $100$, we see that the OH is much closer to the reservoir-only, compared to the two other methods. In case of the IH, the large error by the KBM will be corrected by the entire model, namely the input matrix $\textbf{W}_\text{in}$, the reservoir and the output matrix $\textbf{W}_\text{out}$. Obviously this effect is larger than $\textbf{W}_\text{out}$. This seems to mitigate more the influence of the large error KBM, compare to the OH. The FH inherits this advantage of the IH and is thus more robust than the OH.

\begin{figure}
\includegraphics[width=0.95\linewidth]{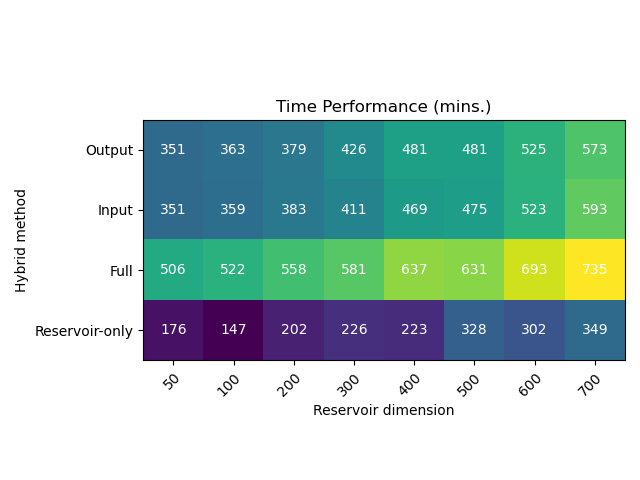}
\caption{Elapsed time for one training and prediction for the ensemble experiment of Fig.\ref{reservoir_rdim}}
\label{time_performance}
\end{figure}

The elapsed time for a training plus a prediction for each case in Fig.\ref{reservoir_rdim} is shown in Fig.\ref{time_performance}. The computation has been done by the same CPU previously mentioned (12th Gen Intel(R) Core(TM) i9-12900K). Each case has been executed by a core. No multi-core computing has been used. The workstation containing the above CPUs were shared with other users. For this reason, one should take into consideration that the following elapsed time measurement has not been done in an ideal environment to measure the computation speed. One can see a tendency such that the elapsed time is in the order of $\mathbf{Reservoir-only}<\mathbf{OH}\approx\mathbf{IH}<\mathbf{FH}$. Without having the extra computation related to the hybrid method, the Reservoir-only is the fastest, as expected. the OH and the IH do not show a significant difference. The FH is the slowest due to its heavy computation for its inner IH as well as the OH. Also as expected, the reservoir dimension $r_\text{dim}$ is clearly the major factor to increase the computation time. These observations indicate that from the point of view of CPU needs it is a better choice to use the OH or the IH with a smaller reservoir dimension $r_\text{dim}$.

From the above observations, one can say that the OH or the FH of a small reservoir dimension such as $r_\text{dim}=50, 300, 100$ is the best choice for this use-case, when the model error $e$ is small. On the other hand, the IH or FH with higher dimensions such as $r_\text{dim}=400$ or more tend to perform better than the OH. It should also be noticed that a higher reservoir dimension takes more time for the training and the prediction. Further, the FH is more time-consuming compared to the IH and OH due to its more complex and larger structure.

\begin{figure}[h!]
  \begin{minipage}[b]{\linewidth}
  \includegraphics[width=\linewidth]{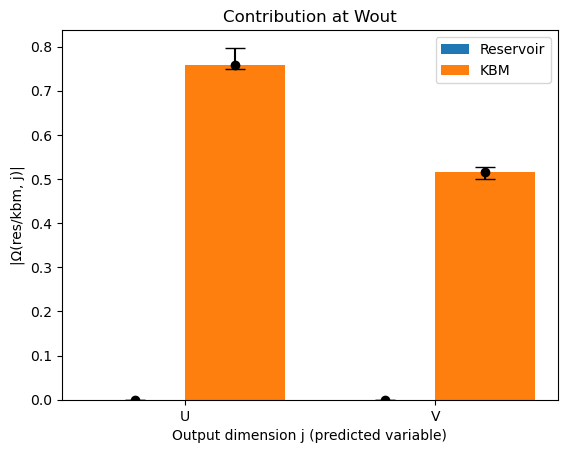}
  \medskip\footnotesize\centering (a) $e=0$
  \end{minipage}
  \vfill  
  \begin{minipage}[b]{\linewidth}
  \includegraphics[width=\linewidth]{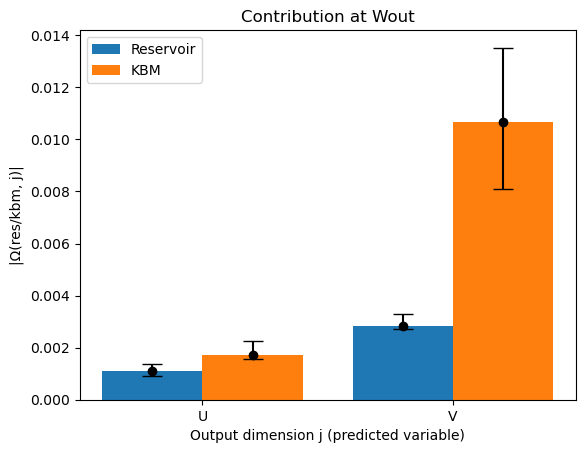}
  \medskip\footnotesize\centering (b) $e=0.1$
  \end{minipage} 
  \caption{Contribution of Reservoir/KBM in $W_{out}$ for no model errors $e=0$ (a) and a model error of $e=0.1$ (b). The contribution is estimated by calculating the median of all the $n_\text{A}\times n_\text{T}$ predictions and displayed with the corresponding standard deviations (lower/higher quartile) across each dimension.}  \label{wout_contribution1}
\end{figure}

\begin{figure}[h!]
  \begin{minipage}[b]{\linewidth}
  \includegraphics[width=\linewidth]{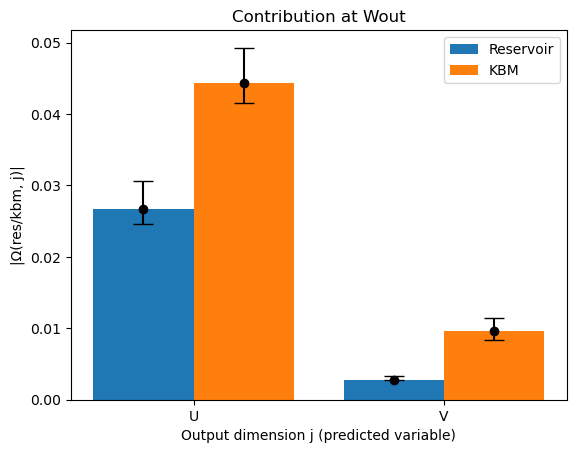}
  \medskip\footnotesize\centering (a) $e=5$ 
  \end{minipage}
  \begin{minipage}[b]{\linewidth}
  \includegraphics[width=\linewidth]{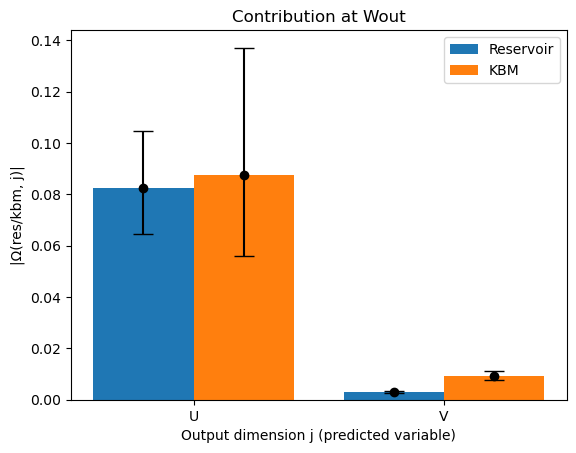}
  \medskip\footnotesize\centering (b) $e=100$
  \end{minipage}    
  \caption{Same as Fig. \ref{wout_contribution1} but for large model errors, i.e. $e=5$ (a) and $e=100$ (b).}  \label{wout_contribution2}
\end{figure}

Fig.\ref{wout_contribution1} and \ref{wout_contribution2} visualize the ratio of the contribution (Reservoir/KBM) in the output matrix $\mathbf{{W}}_\text{out}$ in the OH. On the one hand, we want to check the correct functioning of the OH setup. On the other hand, we want to investigate how the contributions from the (data-driven) RC part and the KBM part are distributed - especially when the model error is increased. 
The model error was set to $e=0, 0.1, 5, 100$, respectively. 
For $e=0$, the noise ratio $\alpha$ and the regularization term $\beta$ in the training data set were set to $0$ to check if the KBM part works correctly as designed. The ensemble experiment has been performed with 20 experiments in total. We followed the method described in Sec.\ref{sec:ensemble} with 5 training-sections and no prediction-sections, ($n_\text{T}=5$ and $n_\text{P}=0$). The adjacency matrix $A$ was initialized 4 times randomly ($n_\text{A}=4$) and 5 training sections were performed in each ${A}$ ($n_\text{T}=5$). During the 5 consecutive training sections, $A$ was kept constant. With $e=0$, the trained model refers completely to the side of the KBM, as expected. In the prediction of $U$ the contribution of the reservoir part increases compared to the KBM as the error increases. In the prediction of $V$, this tendency is much less pronounced and the KBM contribution is still larger than the one from the reservoir even at $e=100$. This is because $V$ does not have the error $\epsilon$ in the KBM.
The increase of the reservoir contribution with the increase of the error in the model confirms the expected functioning of the OH. It is remarkable that for the Barkley model the contributions from the KBM remain larger than the one from RC even for very large model errors. This means that even very inaccurate models do significantly contribute to the prediction and are thus worth being included in a hybrid RC approach.

\section{Conclusions and Outlook}
\label{sec:conclusions}
We developed a framework for hybrid Reservoir Computing supported by Local States for high dimensional systems. The predictions by different methods (non-hybrid, input-hybrid, output-hybrid and full-hybrid) have been conducted with the Barkley model simulating non-linear chaotic excitable medium $80\times80\times2=12,800$ dimensions. The performance of each method were investigated. RC successfully predicted the Barkley model with and without the hybrid approach. The subsequent analyses showed that all hybrid methods significantly improve the prediction performance. The OH and the FH of a small reservoir dimension such as $r_\text{dim}=50, 100$ is the best choice for this use case, when the model error $e$ is small. Given the smaller CPU needs for OH and especially the better interpretability of it, OH is to be favored.
On the other hand, the IH or FH with higher dimensions such as $r_\text{dim}=400$ or more tend to perform better than the OH. 
It was also shown that a higher reservoir dimension takes more time for the training and the prediction. The FH setup is also more time-consuming as compared to the IH and OH. Generally, it may thus be advisable to test the three setups for a given application and  select the best suited one that optimizes between counteracting
factors of prediction performance and CPU requirements.\\
In summary, the combination of locals sates and hybrid RC  expands the application potential of hybrid RC to the class of high-dimensional, spatially extended complex systems. Conceivable application that we will consider in the future are time evolutionary non-linear systems that are represented in the form of two- or three-dimensional image, such as flow simulations in fluid dynamics and atmospheric dynamics as well as excitable media.

\section*{Acknowledgement}
We thank R. Zimmermann and U. Parlitz for providing us access to the simulation code of the Barkeley model.

\section*{Data Availability Statement}
The data that support the findings of this study are available from the corresponding author upon reasonable request.



\section*{References}\label{sec:references}
\bibliography{high_dim_hybrid_rc}

\providecommand{\noopsort}[1]{}\providecommand{\singleletter}[1]{#1}%
\begin{thebibliography}{21}%
\makeatletter
\providecommand \@ifxundefined [1]{%
 \@ifx{#1\undefined}
}%
\providecommand \@ifnum [1]{%
 \ifnum #1\expandafter \@firstoftwo
 \else \expandafter \@secondoftwo
 \fi
}%
\providecommand \@ifx [1]{%
 \ifx #1\expandafter \@firstoftwo
 \else \expandafter \@secondoftwo
 \fi
}%
\providecommand \natexlab [1]{#1}%
\providecommand \enquote  [1]{``#1''}%
\providecommand \bibnamefont  [1]{#1}%
\providecommand \bibfnamefont [1]{#1}%
\providecommand \citenamefont [1]{#1}%
\providecommand \href@noop [0]{\@secondoftwo}%
\providecommand \href [0]{\begingroup \@sanitize@url \@href}%
\providecommand \@href[1]{\@@startlink{#1}\@@href}%
\providecommand \@@href[1]{\endgroup#1\@@endlink}%
\providecommand \@sanitize@url [0]{\catcode `\\12\catcode `\$12\catcode `\&12\catcode `\#12\catcode `\^12\catcode `\_12\catcode `\%12\relax}%
\providecommand \@@startlink[1]{}%
\providecommand \@@endlink[0]{}%
\providecommand \url  [0]{\begingroup\@sanitize@url \@url }%
\providecommand \@url [1]{\endgroup\@href {#1}{\urlprefix }}%
\providecommand \urlprefix  [0]{URL }%
\providecommand \Eprint [0]{\href }%
\providecommand \doibase [0]{http://dx.doi.org/}%
\providecommand \selectlanguage [0]{\@gobble}%
\providecommand \bibinfo  [0]{\@secondoftwo}%
\providecommand \bibfield  [0]{\@secondoftwo}%
\providecommand \translation [1]{[#1]}%
\providecommand \BibitemOpen [0]{}%
\providecommand \bibitemStop [0]{}%
\providecommand \bibitemNoStop [0]{.\EOS\space}%
\providecommand \EOS [0]{\spacefactor3000\relax}%
\providecommand \BibitemShut  [1]{\csname bibitem#1\endcsname}%
\let\auto@bib@innerbib\@empty
\bibitem [{\citenamefont {Jaeger}(2001)}]{jaeger2001echo}%
  \BibitemOpen
  \bibfield  {author} {\bibinfo {author} {\bibfnamefont {H.}~\bibnamefont {Jaeger}},\ }\bibfield  {title} {\enquote {\bibinfo {title} {The “echo state” approach to analysing and training recurrent neural networks-with an erratum note},}\ }\href@noop {} {\bibfield  {journal} {\bibinfo  {journal} {Bonn, Germany: German National Research Center for Information Technology GMD Technical Report}\ }\textbf {\bibinfo {volume} {148}},\ \bibinfo {pages} {13} (\bibinfo {year} {2001})}\BibitemShut {NoStop}%
\bibitem [{\citenamefont {Maass}, \citenamefont {Natschl{\"a}ger},\ and\ \citenamefont {Markram}(2002)}]{maass2002real}%
  \BibitemOpen
  \bibfield  {author} {\bibinfo {author} {\bibfnamefont {W.}~\bibnamefont {Maass}}, \bibinfo {author} {\bibfnamefont {T.}~\bibnamefont {Natschl{\"a}ger}}, \ and\ \bibinfo {author} {\bibfnamefont {H.}~\bibnamefont {Markram}},\ }\bibfield  {title} {\enquote {\bibinfo {title} {Real-time computing without stable states: A new framework for neural computation based on perturbations},}\ }\href@noop {} {\bibfield  {journal} {\bibinfo  {journal} {Neural computation}\ }\textbf {\bibinfo {volume} {14}},\ \bibinfo {pages} {2531--2560} (\bibinfo {year} {2002})}\BibitemShut {NoStop}%
\bibitem [{\citenamefont {Jaeger}\ and\ \citenamefont {Haas}(2004)}]{jaeger2004harnessing}%
  \BibitemOpen
  \bibfield  {author} {\bibinfo {author} {\bibfnamefont {H.}~\bibnamefont {Jaeger}}\ and\ \bibinfo {author} {\bibfnamefont {H.}~\bibnamefont {Haas}},\ }\bibfield  {title} {\enquote {\bibinfo {title} {Harnessing nonlinearity: Predicting chaotic systems and saving energy in wireless communication},}\ }\href@noop {} {\bibfield  {journal} {\bibinfo  {journal} {Science}\ }\textbf {\bibinfo {volume} {304}},\ \bibinfo {pages} {78--80} (\bibinfo {year} {2004})}\BibitemShut {NoStop}%
\bibitem [{\citenamefont {Chattopadhyay}, \citenamefont {Hassanzadeh},\ and\ \citenamefont {Subramanian}(2020)}]{chattopadhyay20}%
  \BibitemOpen
  \bibfield  {author} {\bibinfo {author} {\bibfnamefont {A.}~\bibnamefont {Chattopadhyay}}, \bibinfo {author} {\bibfnamefont {P.}~\bibnamefont {Hassanzadeh}}, \ and\ \bibinfo {author} {\bibfnamefont {D.}~\bibnamefont {Subramanian}},\ }\bibfield  {title} {\enquote {\bibinfo {title} {Data-driven predictions of a multiscale lorenz 96 chaotic system using machine-learning methods: reservoir computing, artificial neural network, and long short-term memory network},}\ }\href {\doibase 10.5194/npg-27-373-2020} {\bibfield  {journal} {\bibinfo  {journal} {Nonlinear Processes in Geophysics}\ }\textbf {\bibinfo {volume} {27}},\ \bibinfo {pages} {373--389} (\bibinfo {year} {2020})}\BibitemShut {NoStop}%
\bibitem [{\citenamefont {Bompas}, \citenamefont {Georgeot},\ and\ \citenamefont {Guéry-Odelin}(2020)}]{bompas20}%
  \BibitemOpen
  \bibfield  {author} {\bibinfo {author} {\bibfnamefont {S.}~\bibnamefont {Bompas}}, \bibinfo {author} {\bibfnamefont {B.}~\bibnamefont {Georgeot}}, \ and\ \bibinfo {author} {\bibfnamefont {D.}~\bibnamefont {Guéry-Odelin}},\ }\bibfield  {title} {\enquote {\bibinfo {title} {{Accuracy of neural networks for the simulation of chaotic dynamics: Precision of training data vs precision of the algorithm}},}\ }\href {\doibase 10.1063/5.0021264} {\bibfield  {journal} {\bibinfo  {journal} {Chaos: An Interdisciplinary Journal of Nonlinear Science}\ }\textbf {\bibinfo {volume} {30}} (\bibinfo {year} {2020}),\ 10.1063/5.0021264},\ \bibinfo {note} {113118},\ \Eprint {http://arxiv.org/abs/https://pubs.aip.org/aip/cha/article-pdf/doi/10.1063/5.0021264/14629142/113118\_1\_online.pdf} {https://pubs.aip.org/aip/cha/article-pdf/doi/10.1063/5.0021264/14629142/113118\_1\_online.pdf} \BibitemShut {NoStop}%
\bibitem [{\citenamefont {Shahi}, \citenamefont {Fenton},\ and\ \citenamefont {Cherry}(2022)}]{shahi22}%
  \BibitemOpen
  \bibfield  {author} {\bibinfo {author} {\bibfnamefont {S.}~\bibnamefont {Shahi}}, \bibinfo {author} {\bibfnamefont {F.~H.}\ \bibnamefont {Fenton}}, \ and\ \bibinfo {author} {\bibfnamefont {E.~M.}\ \bibnamefont {Cherry}},\ }\bibfield  {title} {\enquote {\bibinfo {title} {Prediction of chaotic time series using recurrent neural networks and reservoir computing techniques: A comparative study},}\ }\href {\doibase https://doi.org/10.1016/j.mlwa.2022.100300} {\bibfield  {journal} {\bibinfo  {journal} {Machine Learning with Applications}\ }\textbf {\bibinfo {volume} {8}},\ \bibinfo {pages} {100300} (\bibinfo {year} {2022})}\BibitemShut {NoStop}%
\bibitem [{\citenamefont {Pathak}\ \emph {et~al.}(2018{\natexlab{a}})\citenamefont {Pathak}, \citenamefont {Wikner}, \citenamefont {Fussell}, \citenamefont {Chandra}, \citenamefont {Hunt}, \citenamefont {Girvan},\ and\ \citenamefont {Ott}}]{pathak2018hybrid}%
  \BibitemOpen
  \bibfield  {author} {\bibinfo {author} {\bibfnamefont {J.}~\bibnamefont {Pathak}}, \bibinfo {author} {\bibfnamefont {A.}~\bibnamefont {Wikner}}, \bibinfo {author} {\bibfnamefont {R.}~\bibnamefont {Fussell}}, \bibinfo {author} {\bibfnamefont {S.}~\bibnamefont {Chandra}}, \bibinfo {author} {\bibfnamefont {B.~R.}\ \bibnamefont {Hunt}}, \bibinfo {author} {\bibfnamefont {M.}~\bibnamefont {Girvan}}, \ and\ \bibinfo {author} {\bibfnamefont {E.}~\bibnamefont {Ott}},\ }\bibfield  {title} {\enquote {\bibinfo {title} {Hybrid forecasting of chaotic processes: Using machine learning in conjunction with a knowledge-based model},}\ }\href@noop {} {\bibfield  {journal} {\bibinfo  {journal} {Chaos: An Interdisciplinary Journal of Nonlinear Science}\ }\textbf {\bibinfo {volume} {28}} (\bibinfo {year} {2018}{\natexlab{a}})}\BibitemShut {NoStop}%
\bibitem [{\citenamefont {{Brunton}}, \citenamefont {{Proctor}},\ and\ \citenamefont {{Kutz}}(2016)}]{brunton16}%
  \BibitemOpen
  \bibfield  {author} {\bibinfo {author} {\bibfnamefont {S.~L.}\ \bibnamefont {{Brunton}}}, \bibinfo {author} {\bibfnamefont {J.~L.}\ \bibnamefont {{Proctor}}}, \ and\ \bibinfo {author} {\bibfnamefont {J.~N.}\ \bibnamefont {{Kutz}}},\ }\bibfield  {title} {\enquote {\bibinfo {title} {{Discovering governing equations from data by sparse identification of nonlinear dynamical systems}},}\ }\href {\doibase 10.1073/pnas.1517384113} {\bibfield  {journal} {\bibinfo  {journal} {Proceedings of the National Academy of Science}\ }\textbf {\bibinfo {volume} {113}},\ \bibinfo {pages} {3932--3937} (\bibinfo {year} {2016})},\ \Eprint {http://arxiv.org/abs/1509.03580} {arXiv:1509.03580 [math.DS]} \BibitemShut {NoStop}%
\bibitem [{\citenamefont {{Ma}}\ \emph {et~al.}(2022)\citenamefont {{Ma}}, \citenamefont {{Haluszczynski}}, \citenamefont {{Prosperino}},\ and\ \citenamefont {{R{\"a}th}}}]{ma22}%
  \BibitemOpen
  \bibfield  {author} {\bibinfo {author} {\bibfnamefont {H.}~\bibnamefont {{Ma}}}, \bibinfo {author} {\bibfnamefont {A.}~\bibnamefont {{Haluszczynski}}}, \bibinfo {author} {\bibfnamefont {D.}~\bibnamefont {{Prosperino}}}, \ and\ \bibinfo {author} {\bibfnamefont {C.}~\bibnamefont {{R{\"a}th}}},\ }\bibfield  {title} {\enquote {\bibinfo {title} {{Identifying causality drivers and deriving governing equations of nonlinear complex systems}},}\ }\href {\doibase 10.1063/5.0102250} {\bibfield  {journal} {\bibinfo  {journal} {Chaos}\ }\textbf {\bibinfo {volume} {32}},\ \bibinfo {eid} {103128} (\bibinfo {year} {2022})}\BibitemShut {NoStop}%
\bibitem [{\citenamefont {{K{\"o}ster}}\ \emph {et~al.}(2023)\citenamefont {{K{\"o}ster}}, \citenamefont {{Patel}}, \citenamefont {{Wikner}}, \citenamefont {{Jaurigue}},\ and\ \citenamefont {{L{\"u}dge}}}]{koester23}%
  \BibitemOpen
  \bibfield  {author} {\bibinfo {author} {\bibfnamefont {F.}~\bibnamefont {{K{\"o}ster}}}, \bibinfo {author} {\bibfnamefont {D.}~\bibnamefont {{Patel}}}, \bibinfo {author} {\bibfnamefont {A.}~\bibnamefont {{Wikner}}}, \bibinfo {author} {\bibfnamefont {L.}~\bibnamefont {{Jaurigue}}}, \ and\ \bibinfo {author} {\bibfnamefont {K.}~\bibnamefont {{L{\"u}dge}}},\ }\bibfield  {title} {\enquote {\bibinfo {title} {{Data-informed reservoir computing for efficient time-series prediction}},}\ }\href {\doibase 10.1063/5.0152311} {\bibfield  {journal} {\bibinfo  {journal} {Chaos}\ }\textbf {\bibinfo {volume} {33}},\ \bibinfo {eid} {073109} (\bibinfo {year} {2023})}\BibitemShut {NoStop}%
\bibitem [{\citenamefont {Shahi}\ \emph {et~al.}(2021)\citenamefont {Shahi}, \citenamefont {Marcotte}, \citenamefont {Herndon}, \citenamefont {Fenton}, \citenamefont {Shiferaw},\ and\ \citenamefont {Cherry}}]{shahi2021long}%
  \BibitemOpen
  \bibfield  {author} {\bibinfo {author} {\bibfnamefont {S.}~\bibnamefont {Shahi}}, \bibinfo {author} {\bibfnamefont {C.~D.}\ \bibnamefont {Marcotte}}, \bibinfo {author} {\bibfnamefont {C.~J.}\ \bibnamefont {Herndon}}, \bibinfo {author} {\bibfnamefont {F.~H.}\ \bibnamefont {Fenton}}, \bibinfo {author} {\bibfnamefont {Y.}~\bibnamefont {Shiferaw}}, \ and\ \bibinfo {author} {\bibfnamefont {E.~M.}\ \bibnamefont {Cherry}},\ }\bibfield  {title} {\enquote {\bibinfo {title} {Long-time prediction of arrhythmic cardiac action potentials using recurrent neural networks and reservoir computing},}\ }\href@noop {} {\bibfield  {journal} {\bibinfo  {journal} {Frontiers in physiology}\ }\textbf {\bibinfo {volume} {12}},\ \bibinfo {pages} {734178} (\bibinfo {year} {2021})}\BibitemShut {NoStop}%
\bibitem [{\citenamefont {Doan}, \citenamefont {Polifke},\ and\ \citenamefont {Magri}(2019)}]{doan2019physics}%
  \BibitemOpen
  \bibfield  {author} {\bibinfo {author} {\bibfnamefont {N.~A.~K.}\ \bibnamefont {Doan}}, \bibinfo {author} {\bibfnamefont {W.}~\bibnamefont {Polifke}}, \ and\ \bibinfo {author} {\bibfnamefont {L.}~\bibnamefont {Magri}},\ }\bibfield  {title} {\enquote {\bibinfo {title} {Physics-informed echo state networks for chaotic systems forecasting},}\ }in\ \href@noop {} {\emph {\bibinfo {booktitle} {Computational Science--ICCS 2019: 19th International Conference, Faro, Portugal, June 12--14, 2019, Proceedings, Part IV 19}}}\ (\bibinfo {organization} {Springer},\ \bibinfo {year} {2019})\ pp.\ \bibinfo {pages} {192--198}\BibitemShut {NoStop}%
\bibitem [{\citenamefont {Doan}, \citenamefont {Polifke},\ and\ \citenamefont {Magri}(2021)}]{doan2021short}%
  \BibitemOpen
  \bibfield  {author} {\bibinfo {author} {\bibfnamefont {N.~A.~K.}\ \bibnamefont {Doan}}, \bibinfo {author} {\bibfnamefont {W.}~\bibnamefont {Polifke}}, \ and\ \bibinfo {author} {\bibfnamefont {L.}~\bibnamefont {Magri}},\ }\bibfield  {title} {\enquote {\bibinfo {title} {Short-and long-term predictions of chaotic flows and extreme events: a physics-constrained reservoir computing approach},}\ }\href@noop {} {\bibfield  {journal} {\bibinfo  {journal} {Proceedings of the Royal Society A}\ }\textbf {\bibinfo {volume} {477}},\ \bibinfo {pages} {20210135} (\bibinfo {year} {2021})}\BibitemShut {NoStop}%
\bibitem [{\citenamefont {Duncan}\ and\ \citenamefont {R{\"a}th}(2023)}]{duncan2023optimizing}%
  \BibitemOpen
  \bibfield  {author} {\bibinfo {author} {\bibfnamefont {D.}~\bibnamefont {Duncan}}\ and\ \bibinfo {author} {\bibfnamefont {C.}~\bibnamefont {R{\"a}th}},\ }\bibfield  {title} {\enquote {\bibinfo {title} {Optimizing the combination of data-driven and model-based elements in hybrid reservoir computing},}\ }\href@noop {} {\bibfield  {journal} {\bibinfo  {journal} {Chaos: An Interdisciplinary Journal of Nonlinear Science}\ }\textbf {\bibinfo {volume} {33}} (\bibinfo {year} {2023})}\BibitemShut {NoStop}%
\bibitem [{\citenamefont {Parlitz}\ and\ \citenamefont {Merkwirth}(2000)}]{parlitz2000prediction}%
  \BibitemOpen
  \bibfield  {author} {\bibinfo {author} {\bibfnamefont {U.}~\bibnamefont {Parlitz}}\ and\ \bibinfo {author} {\bibfnamefont {C.}~\bibnamefont {Merkwirth}},\ }\bibfield  {title} {\enquote {\bibinfo {title} {Prediction of spatiotemporal time series based on reconstructed local states},}\ }\href@noop {} {\bibfield  {journal} {\bibinfo  {journal} {Physical review letters}\ }\textbf {\bibinfo {volume} {84}},\ \bibinfo {pages} {1890} (\bibinfo {year} {2000})}\BibitemShut {NoStop}%
\bibitem [{\citenamefont {Pathak}\ \emph {et~al.}(2018{\natexlab{b}})\citenamefont {Pathak}, \citenamefont {Hunt}, \citenamefont {Girvan}, \citenamefont {Lu},\ and\ \citenamefont {Ott}}]{pathak2018model}%
  \BibitemOpen
  \bibfield  {author} {\bibinfo {author} {\bibfnamefont {J.}~\bibnamefont {Pathak}}, \bibinfo {author} {\bibfnamefont {B.}~\bibnamefont {Hunt}}, \bibinfo {author} {\bibfnamefont {M.}~\bibnamefont {Girvan}}, \bibinfo {author} {\bibfnamefont {Z.}~\bibnamefont {Lu}}, \ and\ \bibinfo {author} {\bibfnamefont {E.}~\bibnamefont {Ott}},\ }\bibfield  {title} {\enquote {\bibinfo {title} {Model-free prediction of large spatiotemporally chaotic systems from data: A reservoir computing approach},}\ }\href@noop {} {\bibfield  {journal} {\bibinfo  {journal} {Physical review letters}\ }\textbf {\bibinfo {volume} {120}},\ \bibinfo {pages} {024102} (\bibinfo {year} {2018}{\natexlab{b}})}\BibitemShut {NoStop}%
\bibitem [{\citenamefont {Zimmermann}\ and\ \citenamefont {Parlitz}(2018)}]{zimmermann2018observing}%
  \BibitemOpen
  \bibfield  {author} {\bibinfo {author} {\bibfnamefont {R.~S.}\ \bibnamefont {Zimmermann}}\ and\ \bibinfo {author} {\bibfnamefont {U.}~\bibnamefont {Parlitz}},\ }\bibfield  {title} {\enquote {\bibinfo {title} {Observing spatio-temporal dynamics of excitable media using reservoir computing},}\ }\href@noop {} {\bibfield  {journal} {\bibinfo  {journal} {Chaos: An Interdisciplinary Journal of Nonlinear Science}\ }\textbf {\bibinfo {volume} {28}} (\bibinfo {year} {2018})}\BibitemShut {NoStop}%
\bibitem [{\citenamefont {Wikner}\ \emph {et~al.}(2020)\citenamefont {Wikner}, \citenamefont {Pathak}, \citenamefont {Hunt}, \citenamefont {Girvan}, \citenamefont {Arcomano}, \citenamefont {Szunyogh}, \citenamefont {Pomerance},\ and\ \citenamefont {Ott}}]{wikner2020combining}%
  \BibitemOpen
  \bibfield  {author} {\bibinfo {author} {\bibfnamefont {A.}~\bibnamefont {Wikner}}, \bibinfo {author} {\bibfnamefont {J.}~\bibnamefont {Pathak}}, \bibinfo {author} {\bibfnamefont {B.}~\bibnamefont {Hunt}}, \bibinfo {author} {\bibfnamefont {M.}~\bibnamefont {Girvan}}, \bibinfo {author} {\bibfnamefont {T.}~\bibnamefont {Arcomano}}, \bibinfo {author} {\bibfnamefont {I.}~\bibnamefont {Szunyogh}}, \bibinfo {author} {\bibfnamefont {A.}~\bibnamefont {Pomerance}}, \ and\ \bibinfo {author} {\bibfnamefont {E.}~\bibnamefont {Ott}},\ }\bibfield  {title} {\enquote {\bibinfo {title} {Combining machine learning with knowledge-based modeling for scalable forecasting and subgrid-scale closure of large, complex, spatiotemporal systems},}\ }\href@noop {} {\bibfield  {journal} {\bibinfo  {journal} {Chaos: An Interdisciplinary Journal of Nonlinear Science}\ }\textbf {\bibinfo {volume} {30}} (\bibinfo {year} {2020})}\BibitemShut {NoStop}%
\bibitem [{zim()}]{zimgit}%
  \BibitemOpen
  \href@noop {} {\enquote {\bibinfo {title} {Zimmermann's source code of the barkley model},}\ }\bibinfo {howpublished} {\url{https://github.com/zimmerrol/rcp_spatio_temporal}},\ \bibinfo {note} {accessed: 2024-03-25}\BibitemShut {NoStop}%
\bibitem [{\citenamefont {Baur}\ and\ \citenamefont {R\"ath}(2021)}]{baur21}%
  \BibitemOpen
  \bibfield  {author} {\bibinfo {author} {\bibfnamefont {S.}~\bibnamefont {Baur}}\ and\ \bibinfo {author} {\bibfnamefont {C.}~\bibnamefont {R\"ath}},\ }\bibfield  {title} {\enquote {\bibinfo {title} {Predicting high-dimensional heterogeneous time series employing generalized local states},}\ }\href {\doibase 10.1103/PhysRevResearch.3.023215} {\bibfield  {journal} {\bibinfo  {journal} {Phys. Rev. Res.}\ }\textbf {\bibinfo {volume} {3}},\ \bibinfo {pages} {023215} (\bibinfo {year} {2021})}\BibitemShut {NoStop}%
\bibitem [{\citenamefont {Herteux}\ and\ \citenamefont {Räth}(2020)}]{herteux20}%
  \BibitemOpen
  \bibfield  {author} {\bibinfo {author} {\bibfnamefont {J.}~\bibnamefont {Herteux}}\ and\ \bibinfo {author} {\bibfnamefont {C.}~\bibnamefont {Räth}},\ }\bibfield  {title} {\enquote {\bibinfo {title} {{Breaking symmetries of the reservoir equations in echo state networks}},}\ }\href {\doibase 10.1063/5.0028993} {\bibfield  {journal} {\bibinfo  {journal} {Chaos: An Interdisciplinary Journal of Nonlinear Science}\ }\textbf {\bibinfo {volume} {30}},\ \bibinfo {pages} {123142} (\bibinfo {year} {2020})},\ \Eprint {http://arxiv.org/abs/https://pubs.aip.org/aip/cha/article-pdf/doi/10.1063/5.0028993/14110522/123142\_1\_online.pdf} {https://pubs.aip.org/aip/cha/article-pdf/doi/10.1063/5.0028993/14110522/123142\_1\_online.pdf} \BibitemShut {NoStop}%
\end{thebibliography}%

\end{document}